# A Cure for Pathological Behavior in Games that Use Minimax

Bruce Abramson[1]
Department of Computer Science
Columbia University
New York, N.Y. 10027

## Abstract

*Minimax has long been the standard method of evaluating game tree nodes in game-playing programs. The general assumption underlying these programs is that deepening search improves play. Recent work has shown that this assumption is not always valid; for a large class of games and evaluation functions, deepening search decreases the probability of making a correct move. This phenomena is called game tree pathology.*

*Two structural properties of game trees have been suggested as causes of pathology: independence among the values of sibling nodes, and uniform depth of win nodes. This paper examines the relationship between uniform win depth and pathology.*

*A game-playing program is run using two different evaluation functions. The first recognizes wins only at the bottom level, the second at various levels throughout the tree. The minimax procedure behaves pathologically when the first function is used; the second shows no such pathology. This result constitutes the first experimental evidence linking uniform win depth to pathology. The effect of not recognizing wins until the bottom of the tree on the probability of making a correct decision is also analyzed. This analysis may lead to a general characterization of pathology in terms of win distribution.*

## 1 Introduction

Computer programs that play two-player games generally adhere to the paradigm of heuristic game tree search, the minimax procedure [10]. Minimax strategies have been proven optimal for finite two-person zero-sum games of perfect information [2] (p. 71). Unfortunately, theoretical and implemented games tend to differ in one important aspect: finiteness.[2] In theory, before play begins both players can see the entire game tree, including the actual value of each node. In most implementations, however, the trees are too large to be seen at once, forcing players to move without knowing all possible completions of the game. To account for this discrepancy, minimax has been extended to partial game trees by treating statically evaluated tip nodes like leaves; the tree is searched to some arbitrary depth, all nodes at that depth are evaluated, and the estimated values are minimaxed back up the tree. The appeal of this procedure is obvious - since minimax is an optimal strategy for finite games, estimated minimax should approximate an optimal strategy for infinite games.

Unfortunately, the procedure described does not estimate the minimax value, but rather minimaxes estimated values. In general, the two are not equivalent; computing a function of estimates instead of an estimated function is a cardinal sin of statistics. Statistically sound or not, there is a significant collection of game playing programs that attests that not only does minimaxing estimates work, but the deeper the search, (and thus the greater the functional dependence on those estimates), the better the quality of play [7] [1]. Nevertheless, a theoretical difficulty with minimaxing estimates was pointed out in [5]: for a large class of game trees and evaluation functions, as long as the search does not reach the end of the tree, (in which case a correct decision would be guaranteed), searching deeper causes decisions to become increasingly random. The prediction of games exhibiting this type of pathological behavior suggests two interesting questions: Do any known games belong to this class? And why hasn't pathology been observed in existing game playing programs?

---

[2] Technically, the distinction is not finite vs. infinite, but decidable vs. effectively computable. Minimax is a decision procedure which works on all finite trees, regardless of size. Since computers can only store a relatively small number of nodes, however, the minimax value is not effectively computable for most games.

[1] This research was supported in part by the Defense Advanced Research Projects Agency under contract N00039-84-C-0165, and by the National Science Foundation under grant IST-84-18879.

Section 2 discusses work that has been done on board splitting, a game which behaved pathologically when a reasonably accurate evaluation function was used. Some of the structural differences between board splitting and popular nonpathological games are considered as possible causes of pathology. Section 3 identifies one such difference in the tree of the pathological game described in section 2. This structural flaw increases the probability of making an error as search deepens. A new evaluation function for the board splitting game is introduced to correct this flaw. Using this new function, the pathology disappears. Section 4 offers some conclusions and directions for future work.

## 2 A Pathological Game: Board Splitting

Board splitting was devised by Pearl as an example of a game whose tree has a uniform branching factor (B), a uniform leaf depth (D), and randomly distributed wins and losses among the leaves. Play proceeds as follows: a square ($B^D$-by-$B^D$) board is covered with randomly distributed 1's and 0's. The first player splits the board vertically into B sections, keeps one in play, and discards the rest. The second player splits the remaining portion horizontally, doing the same. After D rounds, only one square remains. If that square contains a 1, the horizontal splitter (H) wins. Otherwise, the vertical splitter (V) wins. To compensate V for going first, the board is set up by flipping a coin weighted in her favor, such that a 1 is generated with probability $p < .5$, and a 0 with probability $(1-p) > .5$. The value of p needed to make the game fair is dependent on B [3].[3] In order to use board splitting as a model for more complex games, the tree must be treated as if it were too large to search entirely. The minimax procedure searches the tree to some arbitrary depth, k, where $0 \leq k \leq 2(D-1)$. An heuristic evaluation function is then applied to all nodes at the specified level, and these estimates are minimaxed back up the tree. The search depth is bounded by $2(D-1)$ to insure that neither player can see the last round prematurely. A simple function which has been used in the past assigns each tip node a value equal to the number of 1's it contains [8] [3]. Call this evaluation function N(g). V tries to minimize 1's (thereby maximizing 0's), and H tries to maximize 1's.

$$N(g) = \begin{cases} \text{The number of 1's in } g \\ \quad \text{if } g \text{ is a tip node} \\ \text{MIN}\{N(g') \mid g' \text{ is a child of } g\} \\ \quad \text{if } g \text{ is a min node} \\ \text{MAX}\{N(g') \mid g' \text{ is a child of } g\} \\ \quad \text{if } g \text{ is a max node} \end{cases}$$

Nau showed that N(g) evaluates a given board fairly accurately; the more 1's a board contains, the more likely it is to be a win for H, and the smaller the board, the more accurate the evaluation. Nevertheless, programs that use N(g) behave pathologically for sufficiently large B and D. In other words, the probability of making a correct decision at a given node is not a monotonically nondecreasing function of search depth; there are cases where searching ahead another round (increasing k by 2) decreases the probability of making a correct decision [3] [4]. This result runs counter to the intuition developed through observing other game programs, in which increased search depth improved play, and constitutes an example of the theoretically predicted pathology.

Various cures have been offered for this pathological behavior. Most of them diagnose the minimax procedure as the primary cause, and alleviate pathology by removing minimax. In its place, product propagation rules that estimate the conditional probability of winning the entire game from each node are used [4] [6] [11]. Although this approach has cured all observed pathologies, it has not answered the basic question: why is the minimax procedure nonpathological in games such as chess and checkers? Two (not necessarily contradictory) conjectures have been forwarded, both focusing on an evaluation function's sensitivity to certain characteristics of a game tree's structure. The first, developed by Nau [3] [4], shows that game trees satisfying certain preconditions exhibit pathologies. Among these preconditions is that the value of a tip node may be dependent only to a limited extent on the values of its siblings. Thus, games like chess and checkers, which clearly do not exhibit independence for most standard evaluation functions, are nonpathological. This sibling independence, it has been hypothesized, is the cause of pathology.

Pearl [9] showed that pathology can only be avoided by using evaluation functions whose accuracy improve by over 50% at each successive level in the tree. Most common game trees are not uniform in structure. Rather, they are riddled with early terminal positions, or *traps*. The estimated values of the ancestors of traps are more reliable than those of other nodes at the same level. Although most evaluation functions are not 50% more accurate for a given node at level k+1 than for a given node at level k, the presence of terminal positions in the vicinity of the search frontier significantly improves the function's accuracy when taken over all nodes at the deeper level. Since deeper searches expose more traps, the noise introduced by an additional minimax operation is counterbalanced by increasingly accurate evaluations. This led to the second conjecture: pathology is caused by the absence of traps. According to this hypothesis, the introduction of even a small number of traps may significantly dampen the noise amplification due to minimaxing. [9]. A useful evaluation function, then, should not only discriminate among nodes based on strength, but detect traps as well.

---

[3] Set p equal to the unique solution to the equation $(1-x)^B = x$ in the interval (0,1).

## 3 Understanding and Curing Pathology

Unlike most games, board splitting has a uniform game tree - all leaves are located at level 2D. The concentration of leaves at the bottom of the tree seems to imply an absence of traps. If traps are needed to avoid pathology, board splitting should be pathological regardless of the evaluation function used. However, the salient feature of traps is not that they are leaves, but that the values associated with them are exact, not estimated. Leaves are not the only nodes with this property. Any node that is recognized as a forced win or loss has an exact value associated with it, as well. Thus, the existence of leaves in mid-tree is not crucial to the avoidance of pathology; the recognition of forced wins can serve the same purpose. There are configurations in board splitting which can serve as traps. The most obvious forced wins are boards which contain a row of 1's (win for H), a column of 0's (win for V), or a main diagonal of 1's (win for H, who always goes last). Although a reasonable case could be made in favor of including other patterns, this decision should not affect the basic result: *evaluation functions that recognize forced wins as traps avoid pathology*.

The evaluation function described in section 2, $N(g)$, recognizes wins only at the leaves. Thus, it frequently overlooks forced wins in favor of configurations with more 1's, albeit less strategically arranged. $Y(g)$, shown below, modifies $N(g)$ so that the patterns described above are recognized as forced wins. Tip nodes are evaluated by checking for a row or diagonal of 1's. If such a row exists, the node is assigned the maximum value of $N(g)$, $B^{2D}$ (the number of squares in the initial board). This assures that wherever possible, a forced win will be chosen by H and avoided by V. If a column of 0's exists, the value $-B^{2D}$ does the reverse, guaranteeing that V will choose it and H avoid it. Otherwise, the number of 1's is counted, just like in $N(g)$. These values are then minimaxed back up the tree.

$$E(g) = \begin{cases} B^{2D} & \text{if } g \text{ contains a row or diagonal of 1's} \\ -B^{2D} & \text{if } g \text{ contains a column of 0's} \\ \text{The number of 1's in } g & \text{Otherwise} \end{cases}$$

$$Y(g) = \begin{cases} E(g) & \text{if } g \text{ is a tip node} \\ \min\{N(g') \mid g' \text{ is a child of } g\} & \text{if } g \text{ is a min node} \\ \max\{N(g') \mid g' \text{ is a child of } g\} & \text{if } g \text{ is a max node} \end{cases}$$

### 3.1 Theoretical Predictions

$Y(g)$ differs from $N(g)$ in only one respect: it introduces nodes with completely accurate values in mid-tree. How often will this correct a mistake that $N(g)$ would make? Define an incorrect decision as the selection of a non-trap node as the best (max or min) child of a given parent despite the existence of a trap.[4] Clearly, $N(g)$ and $Y(g)$ will choose the same child of any parent with no traps among its children. If there is a trap, $Y(g)$ will always (correctly) choose it. $N(g)$, which does not look for traps, may or may not. A simplified model can be constructed to determine the effect of increased search depth on the probability that an evaluation function that does not look for traps will find them.

---

[4]Other reasonable defintions are possible. Several authors [9] [4] consider a decision incorrect only if a "loss" node was chosen when a "win" was available. The definition used here considers a decision incorrect if a node of unknown exact value is chosen when a "win" trap should have been recognized.

On her first move, H looks ahead k levels in the tree, (k even), and evaluates square boards using $N(g)$. Consider only the probability of missing a single type of trap, say a row of 1's. (Analogous arguments can be applied to all other cases, namely V's lookahead, evaluating rectangular boards, and other trap patterns).

Let $S = B^{D-k/2-1}$ be the number of rows (and columns) in g, where g is a board at level k in the game tree.
Let p represent the probability that a 1 was placed in a given square in the original board.
Then $P = Pr[g \text{ is a trap}] = (1-(1-p^S)^S)$.

By definition, if $N(g)$ made an incorrect decision, there must be some node at level (k-1), G, with a trap child of maximum value among its trap children, $g_t$, and a non-trap child of maximum value among the non-traps, $g_{nt}$, for which $N(g_{nt}) > N(g_t)$. In other words, $N(g)$ errs if G has a child which is a trap, but the node containing the most 1's is a non-trap.

Let $Pr[NTN] = Pr[G\text{'s child with the most 1's is not a trap}]$

$$= 1 - \sum_{z=S}^{S^2} \{S \binom{S(S-1)}{z-S} Bp^z(1-p)^{S^2-z}$$

$$(\sum_{j=1}^{z} \binom{S^2}{j} p^j(1-p)^{S^2-j})^{B-1}\}$$

Let $Pr[I] = Pr[G\text{'s children include exactly i traps}]$

$$= \binom{B}{i} P^i(1-P)^{B-i}$$

Then $Pr[N(g) \text{ errs}] = \sum_{i=1}^{B-1} Pr[NTN]Pr[I]$

The complexity of $Pr[NTN]$ is due to the unequal amount of useful information about the nodes. By the definitions of traps and non-traps, $S \leq N(g_t) \leq S^2$ and $0 \leq N(g_{nt}) \leq S^2 - S$. Since $N(g_t)$ has a nontrivial lower bound and $N(g_{nt})$ does not, $g_t$ is more likely to contain the maximum number of 1's than $g_{nt}$, and $N(g)$ is a priori more likely to choose a trap than a non-trap. To simplify the analysis, define another evaluation function, $R(g)$, which chooses nodes arbitrarily. The interesting feature of $R(g)$ is that it can be used to calculate the probability of choosing a non-trap as a function of the probability that a given board is a trap.

$Pr[NTRI] = Pr[R \text{ chooses non-trap}|G \text{ has i trap children}] = \frac{B-i}{B}$

Then, $Pr[R(g) \text{ errs}] = \sum_{i=1}^{B-1} Pr[I]Pr[NTRI]$

$$= \sum_{i=1}^{B-1} (\frac{B-i}{B})\binom{B}{i}P^i(1-P)^{B-i}$$

$$= (1-P)-(1-P)^B$$

This term represents $P_k$, the probability that a trap was missed at depth k. To affect the performance of $R(g)$, this error must be propagated back up k levels, and affect the decision made on H's first move. Depth (k-1) is a min level - for a mistake to appear there, it must have been made on all B children of the min node in question. Thus, $P_{k-1} = P_k^B$. Depth (k-2) is a max level, so $P_{k-2}=(1-P_{k-1})-(1-P_{k-1})^B$. This sequence continues alternating as the error propagates upward. At the level of the original decision, $P_0=(1-P_1)-(1-P_1)^B$. $P_0$ can be shown to grow rapidly as k increases. Thus, the probability that $R(g_{nt}) > R(g_t)$ increases as the search deepens. In other words, an evaluation function which does not look for traps becomes decreasingly likely to choose them as search deepens. $N(g)$, like $R(g)$, is such an evaluation function.

The pathological behavior of $N(g)$, then, can be explained as follows: there are a group of nodes in mid-tree which should be recognized as forced wins. These nodes, when they exist, always represent the maximum children of their parent, and should always be chosen (by a MAX operation). As search depth increases, an evaluation function that does not identify these nodes becomes decreasingly likely to choose one of them. The slight edge that $N(g)$ gives $g_t$ over $g_{nt}$ for having at least S 1's will not counter the rapid growth of $P_0$. Thus, $N(g)$, like $R(g)$, can be expected to become less reliable at each successive level searched. $Y(g)$, on the other hand, should encounter no such difficulty.

3.2 Experimental Results

Game tree pathology is an observed phenomena. Even for board splitting, no definite criterion has been developed for predicting when minimaxing $N(g)$ will behave pathologically. The previous section used Pearl's conjecture that pathology is due to the absence of traps to identify a flaw in $N(g)$, its inability to recognize certain obvious patterns as wins or losses. A new evaluation function, $Y(g)$, recognizes those configurations. The probability that an evaluation function that does not recognize traps will err was shown to be an increasing function of search depth. Y(g), by identifying traps, avoids these errors.

Y(g)'s ability to recognize these patterns indicates that it should outperform N(g); it does not prove that Y(g) is nonpathological. It is altogether conceivable that because Y(g) only recognizes some forced wins it will behave pathologically as well. In fact, because pathology is an *observed* phenomena, it is impossible to *prove* that Y(g), or any evaluation function on any game, for that matter, will never behave pathologically. However, it is possible to construct a series of experiments which show that for several cases for which N(g) behaves pathologically, Y(g) does not.

For a fixed B and D, 100 random games were generated. One player sees only her possible next moves, while the other player looks ahead k moves. K varies by 2's, either from 0 to 2(D-1), or from 1 to (2D-3). If the lookahead length is even, tip nodes are MAX nodes (square boards for H, rectangular boards for V). If odd, the tips are MIN nodes (rectangular for H, square for V). To insure that neither player can see the endgame too early, lookahead is always cut off at H's next-to-last move, level 2(D-1) in the original game tree. For each lookahead length, the same 100 games are played, with both players using the same evaluation function, first N(g), then Y(g). The results of these experiments are shown below for three pair of B and D.

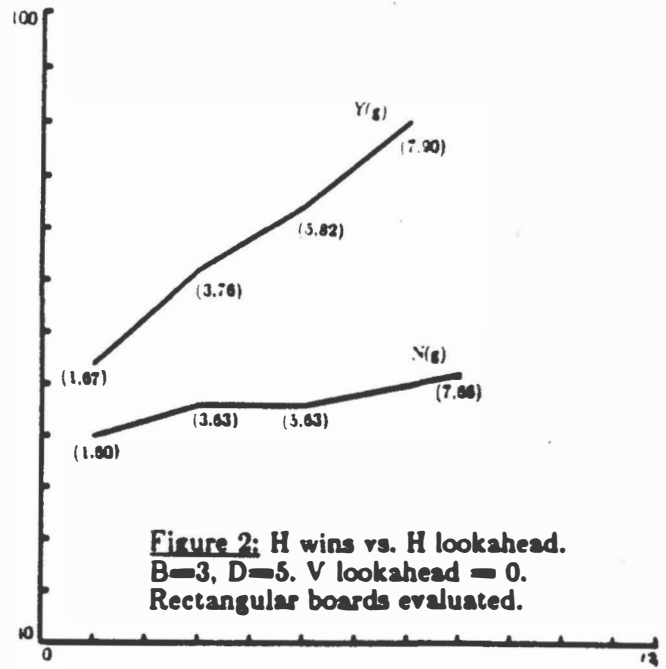

Figure 2: H wins vs. H lookahead. B=3, D=5. V lookahead = 0. Rectangular boards evaluated.

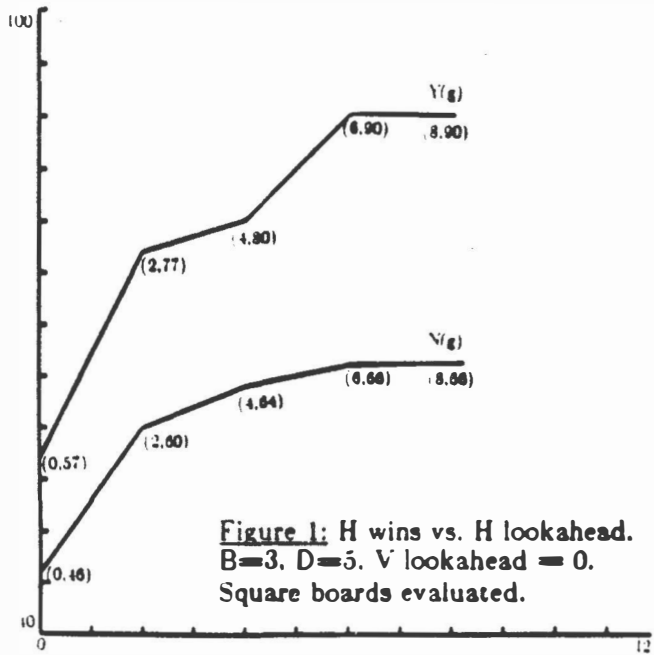

Figure 1: H wins vs. H lookahead. B=3, D=5. V lookahead = 0. Square boards evaluated.

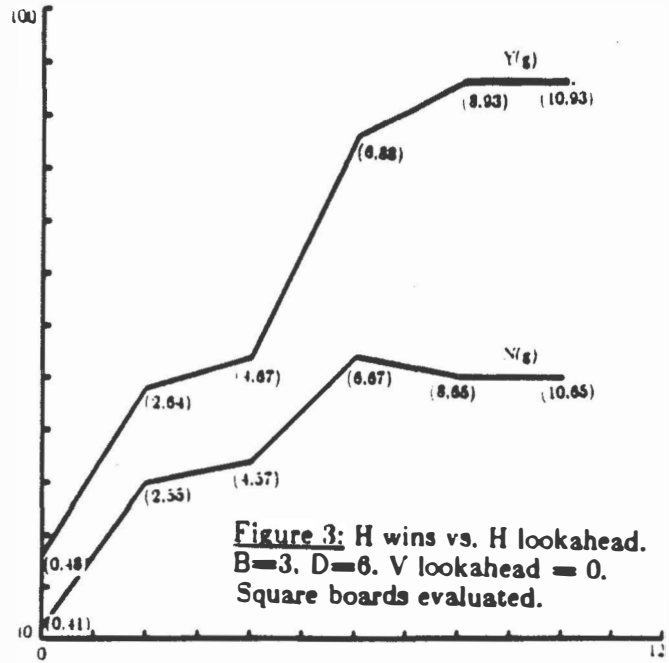

Figure 3: H wins vs. H lookahead. B=3, D=6. V lookahead = 0. Square boards evaluated.

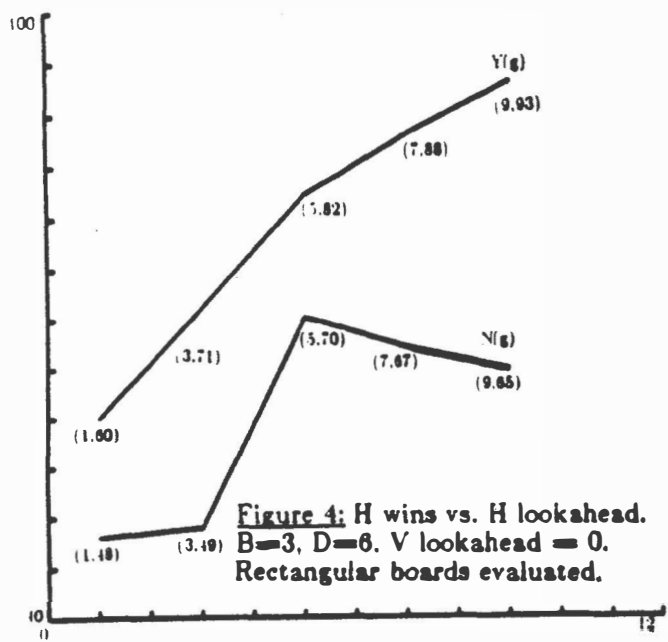

Figure 4: H wins vs. H lookahead. B=3, D=6, V lookahead = 0. Rectangular boards evaluated.

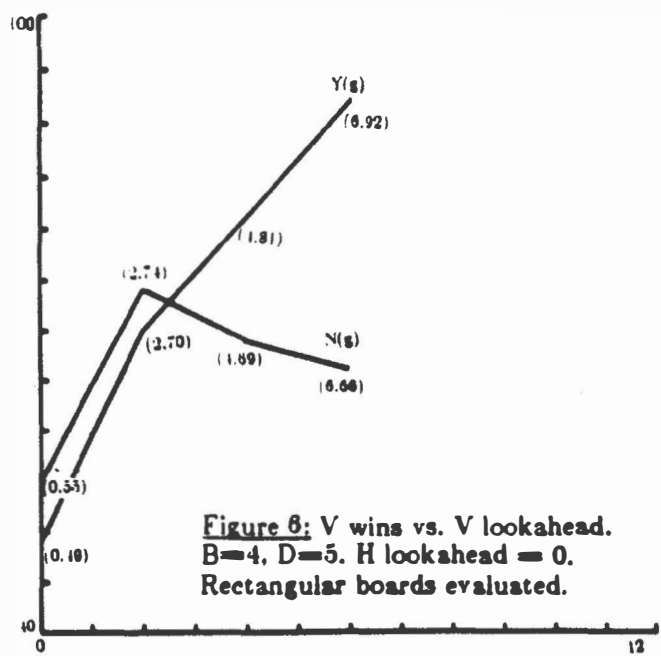

Figure 6: V wins vs. V lookahead. B=4, D=5, H lookahead = 0. Rectangular boards evaluated.

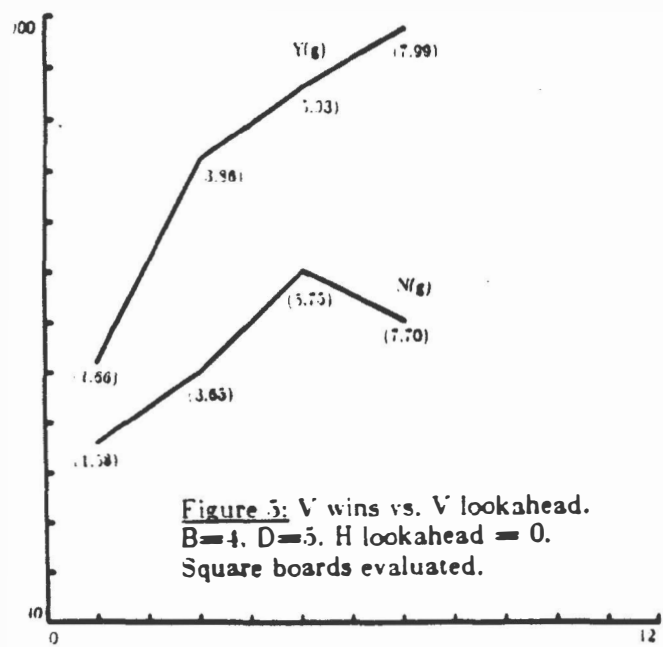

Figure 5: V wins vs. V lookahead. B=4, D=5, H lookahead = 0. Square boards evaluated.

There are several points worth mentioning. First, although p was chosen to make games using $N(g)$ fair [3], the random number generator seems to have favored V - V consistently won over 50% of the games in which both players used $N(g)$ with equal (0 move) lookahead. Second, $Y(g)$ favors H. Since H goes second, diagonals represent wins for H, but not for V. Furthermore, H always looks for wins in smaller boards. In other words, on the $j^{th}$ move, H's wins contain $B^{D-j-1}$ 1's, while V's wins require $B^{D-j}$ 0's. Third, $N(g)$ is not always pathological. Nonpathological behavior is characterized by monotonic nondecreasing functions of victories vs. search depth. For B and D sufficiently small, $N(g)$ behaves nonpathologically (see figures 1 and 2). However, even in these cases, $N(g)$ is rather weak; allowing H to see level 10 in a 12 level tree on her first turn results in only 66 victories in 100 games, as opposed to 90 when $Y(g)$ is used. Finally, and most importantly, in all cases in which $N(g)$ behaved pathologically, $Y(g)$ did not (see figures 3,4,5,6). This constitutes the first empirical evidence supporting Pearl's claim: *introducing traps avoided pathology.*

## 4 Conclusions

The theoretical prediction of game tree pathology in [5], and the subsequent observation of pathological behavior in board splitting [3],

raised an obvious question: What characteristics of game trees cause pathological behavior? Two plausible answers have been posited, independence among sibling nodes [3] [4], and the absence of traps [9]. This paper examined the relationship between traps and pathology in board splitting. The pathological behavior of $N(g)$, an evaluation function which has been shown to evaluate individual boards fairly accurately [3], was explained. Although it performed well on individual boards, $N(g)$'s inability to recognize traps doomed it to frequently missing the best choice. A modified evaluation function, $Y(g)$, was designed to create traps by recognizing certain midgame setups as wins/losses. For several cases in which $N(g)$ behaved pathologically, $Y(g)$ did not. These results offer the first empirical evidence of the importance of traps in avoiding pathology. Furthermore, they extend the definition of traps to include all nodes of known exact value. This extension makes it possible to devise nonpathological evaluation functions for games with uniform structure.

The probabilistic analysis of $R(g)$ outlined in section 3.1 was not dependent on board splitting; a similar argument would hold for any evaluation function failing to recognize traps in the middle of a uniform game tree. Further analyses are now in progress to resolve two major points: the probability with which specific evaluation functions, such as $N(g)$, err, and the exact relationship between $P_0$ and $P$. Expressing $P_0$ as a function of $P$ would give a general formula for the probability of choosing a trap as a function of the probability that a given node is a trap. This, in turn, would give a general characterization of pathology in terms of trap distribution.

## Acknowledgements

I would like to thank Mordechi Yung, Eugene Pinsky, and my advisor, Richard Korf, for their helpful discussions and suggestions.

## References

1. Berliner, H. "The B* Tree Search Algorithm: a Best-first Proof Procedure". *Artificial Intelligence 12* (1979), 23-40.

2. Luce, R.D. and H. Raiffa. *Games and Decisions*. John Wiley and Sons, New York, 1967.

3. Nau, D.S. "An Investigation of the Causes of Pathology in Games". *Artificial Intelligence 19* (1982), 257-278.

4. Nau, D.S. "Pathology on Game Trees Revisited, and an Alternative to Minimaxing". *Artificial Intelligence 21* (1983), 221-244.

5. Nau, D. S. "Decision Quality as a Function of Search Depth on Game Trees". *JACM 30*, 4 (October 1983), 687-708.

6. Nau, D.S., P. Purdom, C.H. Tzeng. Experiments on Alternatives to Minimax. University of Maryland, October, 1983.

7. Nilsson, N.J.. *Principles of Artificial Intelligence.* Tioga, Palo Alto, California, 1980.

8. Pearl, J.. *Heuristics*. Addison-Wesley, Reading, Massachusetts, 1984.

9. Pearl, J. "On the Nature of Pathology in Game Searching". *Artificial Intelligence 20* (1983), 427-453.

10. Shannon, C.E. "Programming a Computer for Playing Chess". *Philosophical Magazine 41* (1950), 256-275.

11. Tzeng, C.H., P.W. Purdom. A Theory of Game Trees. Proceedings of the National Conference on Artificial Intelligence, AAAI, 1983, pp. 416-419.